\documentclass{article} 
\usepackage{iclr2023_conference_tinypaper,times}

\usepackage{amsmath,amsfonts,bm}

\def\eqref#1{equation~\ref{#1}}

\def\1{\bm{1}}

\DeclareMathAlphabet{\mathsfit}{\encodingdefault}{\sfdefault}{m}{sl}
\SetMathAlphabet{\mathsfit}{bold}{\encodingdefault}{\sfdefault}{bx}{n}

\usepackage{hyperref}
\usepackage{listings}
\usepackage{url}
\usepackage{tikz}
\usepackage{tikzsymbols}
\usepackage{xcolor}
\usepackage{pgfplots}
\usepackage{float}
\usepackage{hyperref}
\usepackage[graphicx]{realboxes}
\definecolor{bblue}{HTML}{4F81BD}
\definecolor{rred}{HTML}{C0504D}
\definecolor{ggreen}{HTML}{9BBB59}
\title{An Evaluation Benchmark for Autoformalization in Lean4}

\iclrfinalcopy

\author{Aryan Gulati*, Devanshu Ladsaria*, Shubhra Mishra*, Jasdeep Sidhu*, Brando Miranda  \\
Department of Computer Science \\
Stanford University \\
Stanford, CA 94305 \\
\texttt{\{aryangul, devanshu, shubhra, jasdeep6, brando9\}@cs.stanford.edu} \\
}

\begin{document}

\maketitle
\def\thefootnote{*}\footnotetext{These authors contributed equally to this work}\def\thefootnote{\arabic{footnote}}
\begin{abstract}
Large Language Models (LLMs) hold the potential to revolutionize autoformalization. The introduction of Lean4, a mathematical programming language, presents an unprecedented opportunity to rigorously assess the autoformalization capabilities of LLMs. This paper introduces a novel evaluation benchmark designed for Lean4, applying it to test the abilities of state-of-the-art LLMs, including GPT-3.5, GPT-4, and Gemini Pro. Our comprehensive analysis reveals that, despite recent advancements, these LLMs still exhibit limitations in autoformalization, particularly in more complex areas of mathematics. These findings underscore the need for further development in LLMs to fully harness their potential in scientific research and development. This study not only benchmarks current LLM capabilities but also sets the stage for future enhancements in autoformalization. 
\end{abstract}
\begin{center}
    Benchmark Page: \href{https://huggingface.co/datasets/shubhramishra/autoformalization-benchmark-lean4}{\texttt{HuggingFace}}
\end{center}

\section{Introduction}

Generating formal statements is tedious, but the impressive advances in LLMs’ capabilities show a promising future for autoformalized, verifiable systems \citep{formal-verification-software}. Computer-formalized mathematics has seen advances in many directions, including the rapid development of new computer-interpretable mathematical languages. One such language is Lean4, the non backwards-compatible successor to Lean3. Given the differences between the two languages, a benchmark that evaluates a LLM’s ability to autoformalize into Lean4 has become increasingly important.

\textbf{Contribution}: In this paper, we propose a benchmark of 101 pairs of mathematical formal-informal statements across 17 different topics in math. Then, we manually evaluated three different state of the art LLMs (GPT-3.5, GPT-4, and Gemini Pro) on the benchmark.

Many benchmarks have used the perplexity metric to evaluate autoformalizations \citep{minif2f, azerbayev2023proofnet}. However, this relies on string/pattern matching, which is not a very robust measure of autoformalization, given the fact that LMs may generate correct formalizations that differ in structure or wording. In our paper, we evaluate autoformalizations on a 0-4 scale based on correction effort, as proposed in \citep{mma}. 
Correction effort refers to the amount of necessary adjustments or modifications required to transform the generated formalization output of a LLM into an accurate and fully correct Lean4 formalization. Additionally, we split the statements into math topics, which lets our evaluation extend beyond an accuracy metric, providing a more fine-grained understanding of how LLMs autoformalize, and where more work is still needed.

\section{Methodology and Results}

To assess the autoformalization capabilities of contemporary LLMs, we selected a dataset of 101 theorem statements from mathlib4, a comprehensive library of mathematical theorems formulated in Lean4. The dataset included a wide array of mathematical subjects (Appendix \ref{app:dataset}) ensuring a diverse and representative sample for our analysis. The dataset includes formal statements, their corresponding natural language informalizations, and the specific mathematical topic.

We employed a zero-shot prompting approach with three advanced LLMs: GPT-3.5, GPT-4, and Gemini Pro. This approach involved presenting each model with natural language statements from our dataset and analyzing the formalized outputs they generated (Appendix \ref{app:prompt-output}). We also streamlined the evaluation process by trimming outputs to only include formal Lean statements.

Our evaluation methodology drew inspiration from \citep{mma}, employing a grading scale ranging from 0 to 4. On this scale, a score of 0 indicates a flawless autoformalization, while a score of 4 signifies an output requiring as much correction effort as formalizing a statement from scratch.

Our analysis revealed that the correction efforts for autoformalizations were similar among GPT-3.5 and GPT-4, averaging 2.238. Gemini Pro showed a slightly higher average effort of 2.248. Gemini Pro boasts the most number of autoformalizations with scores of 0s and 1s. However despite this, GPT-4 and Gemini Pro produced more instances with the maximum correction effort of 4 (Appendix \ref{app:graphs}). This is likely because as discussed in \citep{Pichai_Hassabis_2023}, Gemini, with its natively multi-modal design and recent training incorporating Lean4 data, performs better in reasoning tasks. This is a step forward from GPT-4's Mixture of Experts (MoE) design and earlier training phase, which may have had less exposure to Lean4 (as evident from GPT-4's misinterpretation of Lean4 capabilities in Appendix \ref{app:prompt-output}). Both models surpass GPT-3.5, which relies on a monolithic architecture. 

\begin{figure}
    \centering
    \includegraphics[scale=0.41]{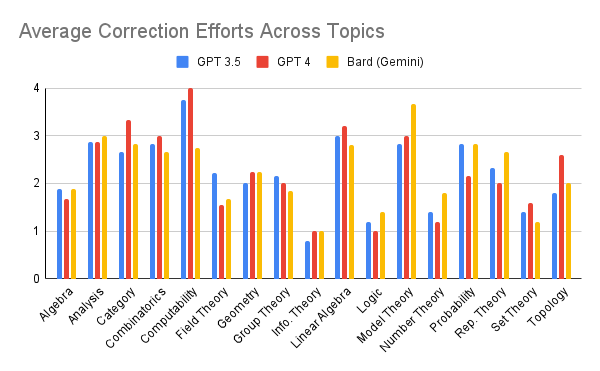}
    \caption{Average Correction Efforts Across Topics}
    \label{fig:correction-topics}
\end{figure}

Figure \ref{fig:correction-topics} reveals performance disparities among LLMs across mathematical subjects, which suggests that the LLMs' performance is subject-dependent. For instance, all LLMs excelled in Information Theory and Logic, but had trouble with category and model theory. We hypothesize that the frequency of these subjects on the internet is related to the performance of the LLM. Another potential reason for the discrepancy between subjects might be attributed to the difficulties of autoformalization. Problems in category theory and model theory are harder to describe even in natural language, so translating it to formal language is a more difficult task in itself. To improve our dataset, we could label the difficulty of each problem statement to correct for correlation between problem- and autoformalization-difficulty. The overall variance suggests that the LLMs' performance is influenced by the subject matter of the theorem, pointing to potential avenues for future research.

\section{Conclusion}

Our research underscores the potential of LLMs in revolutionizing the field of formalization, with implications extending across mathematics, computer science, and engineering. While LLMs can substantially expedite research and development, our findings indicate that even the most sophisticated models currently fall short in achieving accurate autoformalization. This gap highlights the unique opportunity presented by the development of Lean4, serving as a crucial testing ground for enhancing LLM performance in autoformalization and automated theorem proving.

\newpage
\subsubsection*{URM Statement}
We acknowledge that all the authors of this work meets the URM criteria of ICLR 2024 Tiny Papers Track.

\bibliography{iclr2023_conference_tinypaper}
\bibliographystyle{iclr2023_conference_tinypaper}

\appendix
\section{Dataset}
The evaluation benchmark can be accessed via HuggingFace \href{https://huggingface.co/datasets/shubhramishra/autoformalization-benchmark-lean4}{here}.
\section{Inside the Dataset}
\label{app:dataset}
\begin{table}[H]
    \centering
    \begin{tabular}{c|c}
        Topic & Number of problems \\
        \hline
         Algebra & 9 \\
         Analysis & 8 \\
         Category Theory&6 \\
        Combinatorics&6 \\
        Computability&4 \\
        Field Theory&9 \\
        Geometry&8 \\
        Group Theory&6 \\
        Info. Theory&5 \\
        Linear Algebra&5 \\
        Logic&5 \\
        Model Theory&6 \\
        Number Theory&5 \\
        Probability&6 \\
        Rep. Theory&3 \\
        Set Theory&5 \\
        Topology&5 \\
    \end{tabular}
    \caption{The Number of Problems in the Dataset, Based on Topic}
    \label{tab:num-probs}
\end{table}

\section{Sample Prompts and Outputs}
\label{app:prompt-output}
\begin{lstlisting}[mathescape=true]
    $\textbf{Prompt: }$
    "I am providing you with a mathematical statement 
    in natural language. I want you to formalize it in the Lean4 
    language. \n" + natural_language_statement

    $\textbf{Output:}$
    {
    "id": 8,
    "subject": "Information Theory",
    "natural_language_statement": "The Hamming distance of an element to itself is always 0.",
    "ground_truth_formalization": "theorem hammingDist_self (x : $\forall$ i, $\beta$ i) : hammingDist x x = 0 := by%n  rw [hammingDist, card_eq_zero, filter_eq_empty_iff]%n  exact fun _ _ H => H rfl%n#align hamming_dist_self hammingDist_self",
    "gpt_3.5_turbo_response": "import data.list.basic%n%nopen list%n%n-- Define the Hamming distance function%ndef hamming_distance {$\alpha$ : Type} [decidable_eq $\alpha$] (xs ys : list $\alpha$) : $\mathbb{N}$ :=%n  (xs.zip ys).count ($\lambda$ xy, xy.fst \neq xy.snd)%n%n-- Formalize the statement%ntheorem hamming_distance_self_zero {$\alpha$ : Type} [decidable_eq \alpha] (x : $\alpha$) :%n  hamming_distance [x] [x] = 0 :=%nbegin%n  -- Apply the definition of hamming_distance%n  unfold hamming_distance,%n  -- Simplify the expression%n  simp,%nend",
    "gpt_4_response": "In Lean4, the Hamming distance of an element to itself being always 0 can be formalized as follows:%n%n```lean%nimport data.nat.basic -- import necessary library%n%ndef hamming_distance (x y : $\mathbb{N}$) : $\mathbb{N}$ :=%n  nat.popcount (x lxor y) -- definition of Hamming distance%n%ntheorem hamming_self (n : $\mathbb{N}$) : hamming_distance n n = 0 :=%nbegin%n  rw hamming_distance,%n  rw nat.lxor_self,%n  exact nat.popcount_zero,%nend%n```%n%n. This code first imports the necessary library for natural numbers. Then it defines the Hamming distance as the population count (number of 1's) in the binary representation of the bitwise exclusive or (lxor) of two numbers. The theorem `hamming_self` states that the Hamming distance of a number to itself is always 0. The proof is straightforward: the bitwise exclusive or of a number with itself is always 0, and the population count of 0 is 0.",
    "gemini_pro_response": "theorem hamming_distance_to_self_is_zero ($\alpha$ : Type) [decidable_eq \alpha] (x : $\alpha$) :%n  hamming_distance x x = 0 :%n  by refl"
    }
    
    $\textbf{Scores given to Problem 8:}$ GPT-3.5 - 1. GPT-4 - 1. Gemini Pro - 1. All autoformalizations were near perfect with minor changes necessary, which led us to giving scores of 1 for all models.

    $\textbf{Output:}$
    {
      "id": 99,
      "subject": "Category Theory",
      "natural_language_statement": "An object is injective if its preadditive Yoneda functor preserves finite colimits.",
      "ground_truth_formalization": "theorem injective_of_preservesFiniteColimits_preadditiveYonedaObj (J : C)%n    [hP : PreservesFiniteColimits (preadditiveYonedaObj J)] : Injective J :=",
      "gpt_3.5_turbo_response": "import category_theory.category%nimport category_theory.functor%nimport category_theory.limits%n%nopen category_theory%n%nuniverses v u%n%nvariables (C : Type u) [category.{v} C]%n%nstructure injective_object (X : C) : Prop :=%n(preadditive_yoneda_functor_preserves_colimits :%n  $\forall$ (J : Type v) [small_category J] (F : J $\rightarrow$ C),%n  limits.preserves_colimits_of_shape (discrete.functor F) (yoneda.obj X))%n%nNote: This formalization assumes that Lean4 has already imported the necessary libraries for category theory. The statement is formalized using the `injective_object` structure, which takes an object `X` in category `C` as a parameter. The structure contains a property `preadditive_yoneda_functor_preserves_colimits`, which asserts that for any small category `J` and functor `F : J $\rightarrow$ C`, the preadditive Yoneda functor preserves finite colimits. The `limits.preserves_colimits_of_shape` function is used to express this preservation property.",
      "gpt_4_response": "Formalizing this statement in Lean4 requires a deep understanding of category theory and the Lean4 language. Unfortunately, Lean4's standard library does not currently support category theory to the extent required to formalize this statement. %n%nHowever, here is a rough sketch of how you might start to define these concepts in Lean4:%n%n```lean%n-- First, we need to define what a category is%nstructure Category :=%n  (Obj : Type)%n  (Hom : Obj $\rightarrow$ Obj $\rightarrow$ Type)%n  (comp : $\Pi$ {X Y Z : Obj}, Hom X Y $\rightarrow$ Hom Y Z $\rightarrow$ Hom X Z)%n  (id : $\Pi$ X : Obj, Hom X X)%n  (assoc : $\Pi$ {W X Y Z : Obj} (f : Hom W X) (g : Hom X Y) (h : Hom Y Z),%n    comp (comp f g) h = comp f (comp g h))%n  (id_left : $\Pi$ {X Y : Obj} (f : Hom X Y), comp (id X) f = f)%n  (id_right : $\Pi$ {X Y : Obj} (f : Hom X Y), comp f (id Y) = f)%n%n-- Then, we need to define what a functor is%nstructure Functor (C D : Category) :=%n  (onObj : C.Obj $\rightarrow$ D.Obj)%n  (onHom : $\Pi$ {X Y : C.Obj}, C.Hom X Y $\rightarrow$ D.Hom (onObj X) (onObj Y))%n  (functor_id : $\Pi$ {X : C.Obj}, onHom (C.id X) = D.id (onObj X))%n  (functor_comp : $\Pi$ {X Y Z : C.Obj} (f : C.Hom X Y) (g : C.Hom Y Z),%n    onHom (C.comp f g) = D.comp (onHom f) (onHom g))%n%n-- Then, we need to define what a preadditive category is%nstructure Preadditive (C : Category) :=%n  (add : $\Pi$ {X Y : C.Obj}, C.Hom X Y $\rightarrow$ C.Hom X Y $\rightarrow$ C.Hom X Y)%n  (zero : $\Pi$ {X Y : C.Obj}, C.Hom X Y)%n  (neg : $\Pi$ {X Y : C.Obj}, C.Hom X Y $\rightarrow$ C.Hom X Y)%n  (sub : $\Pi$ {X Y : C.Obj}, C.Hom X Y $\rightarrow$ C.Hom X Y $\rightarrow$ C.Hom X Y)%n  -- plus the axioms for an additive group...%n%n-- Then, we need to define what a Yoneda functor is%ndef Yoneda (C : Category) (X : C.Obj) : Functor C (Preadditive C) := sorry%n%n-- Then, we need to define what a colimit is%ndef Colimit (F : Functor C D) : D.Obj := sorry%n%n-- Then, we need to define what it means for a functor to preserve colimits%ndef PreservesColimits (F : Functor C D) : Prop :=%n  $\forall$ (G : Functor C D) (c : Colimit G), Colimit (Functor.comp F G) = F.onObj c%n%n-- Finally, we can define what it means for an object to be injective%ndef Injective (C : Category) (X : C.Obj) : Prop :=%n  PreservesColimits (Yoneda C X)%n```%n%nPlease note that this is a very rough sketch and is not complete or correct Lean4 code. The definitions of `Yoneda`, `Colimit`, and `PreservesColimits` are particularly difficult and would require a lot of additional code.",
      "gemini_pro_response": "def injective [additive] (C : Category) : Prop := $\forall$ (F : Functor C^{op} $\rightarrow$ Ab) (X : C), preserves_finite_colimits F $\leftrightarrow$ preserves_finite_colimits (Yoneda.obj C X) $\cong$ F"
    
    }

    $\textbf{Scores given to Problem 99:}$ GPT-3.5 - 2. GPT-4 - 3. Gemini Pro - 2. GPT 4 avoided a direct autoformalization and proceeded to define other concepts in category theory giving it a score of 3. GPT 3.5 however, attempted a direct autoformalization with Lean4 syntax and provided partially sound logic warranting a score of 2. Bard, similar to GPT 3.5, provided a partially correct direct autoformalization with Lean4 syntax warranting a score of 2. 

\end{lstlisting}
\section{Graphing the Correction Effort for Separate Models}
\label{app:graphs}
\begin{tikzpicture}
    \begin{axis}[
        title = Autoformalizing to Lean4 with GPT 3.5,
        width  = 0.5*\textwidth,
        height = 6cm,
        major x tick style = transparent,
        ybar = 2*\pgflinewidth,
        bar width=14pt,
        ymajorgrids = true,
        ylabel = {Counts},
        xlabel = {Correction Effort},
        symbolic x coords={0,1,2, 3, 4},
        xtick = data,
        scaled y ticks = false,
        enlarge x limits = 0.2,
    ]
        \addplot[style={bblue,fill=bblue,mark=none}]
            coordinates {(0, 4) (1,24)(2,29) (3, 32) (4, 12)};

    \end{axis}
\end{tikzpicture}
\begin{tikzpicture}
    \begin{axis}[
        title = Autoformalizing to Lean4 with GPT 4,
        width  = 0.5*\textwidth,
        height = 6cm,
        major x tick style = transparent,
        ybar = 2*\pgflinewidth,
        bar width=14pt,
        ymajorgrids = true,
        ylabel = {Counts},
        xlabel = {Correction Effort},
        symbolic x coords={0,1,2, 3, 4},
        xtick = data,
        scaled y ticks = false,
        enlarge x limits = 0.2,
        ymin = 0,
    ]
        \addplot[style={rred,fill=rred,mark=none}]
            coordinates {(0, 2) (1,29)(2,30) (3, 23) (4, 17)};
    \end{axis}
\end{tikzpicture} 

\begin{tikzpicture}
    \begin{axis}[
        title = Autoformalizing to Lean4 with Gemini Pro,
        width = 0.5*\textwidth,
        height = 6cm,
        major x tick style = transparent,
        ybar = 2*\pgflinewidth,
        bar width=14pt,
        ymajorgrids = true,
        ylabel = {Counts},
        xlabel = {Correction Effort},
        symbolic x coords={0,1,2, 3, 4},
        xtick = data,
        scaled y ticks = false,
        enlarge x limits = 0.25
    ]
        \addplot[style={ggreen,fill=ggreen,mark=none}]
            coordinates {(0, 3) (1,31)(2,24) (3, 24) (4, 19)};

    \end{axis}
\end{tikzpicture}

\end{document}